\documentclass[letterpaper, 10 pt, conference]{ieeeconf}  

\usepackage{cite}
\usepackage{amsmath,amssymb,mathtools}
\usepackage{stfloats}
\usepackage{etoolbox}
\usepackage{graphicx}
\usepackage{capt-of}
\usepackage[bookmarks=true]{hyperref}
\hypersetup{
    colorlinks=true,
    linkcolor=blue,
    filecolor=blue,      
    urlcolor=blue,
    citecolor=magenta,
}
\usepackage{xurl}
\usepackage{xcolor}
\definecolor{LinkPink}{HTML}{C2185B}
\usepackage{placeins}

\IEEEoverridecommandlockouts                              

\usepackage{xcolor}

\title{\LARGE \bf
StageCraft: Execution Aware Mitigation of Distractor and Obstruction Failures in VLA Models
}

\newcommand{\insertfig}{%
  \includegraphics[width=0.95\linewidth]{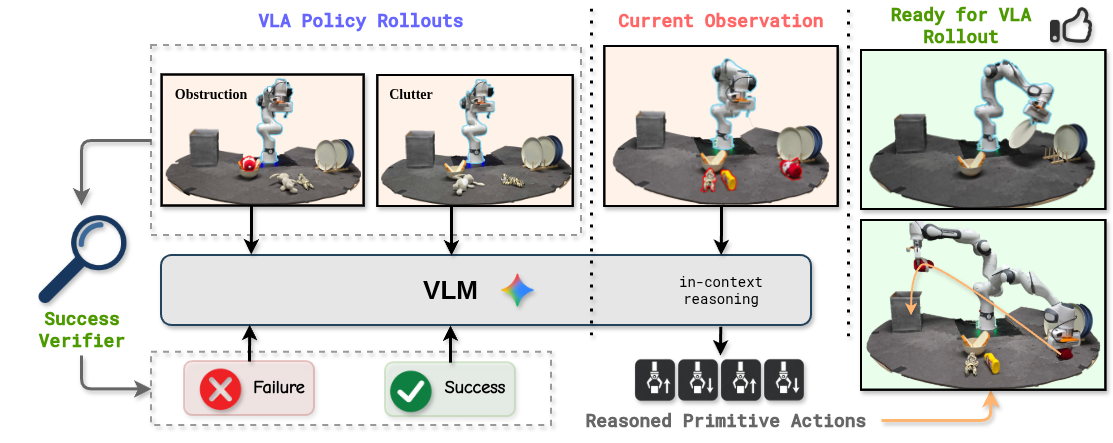} %
  \captionof{figure}{\textit{\textbf{StageCraft}} is a training-free method for improving policy performance that takes as input a history of policy rollouts and their associated terminal rewards, and uses in-context reasoning to infer a sequence of environment modification actions that avoid anticipated failures and increase the policy’s success rate in the presence of distractors and obstructions in the environment.}\label{fig:banner}%
  \vspace{-10pt}
}

\makeatletter
\apptocmd{\@maketitle}{%
  \par\bigskip
  \setcounter{figure}{0}%
  \centering\insertfig\par\bigskip
}{}{}
\makeatother

\author{
Kartikay Milind Pangaonkar$^{*}$, Prabin Kumar Rath$^{*}$, Omkar Patil$^{*}$, Nakul Gopalan\\
Arizona State University\\
\thanks{*Equal contribution}
\thanks{\texttt{\{kpangaon, prath4, opatil3, ng\}@asu.edu}}
}

\begin{document}

\maketitle
\thispagestyle{empty}
\pagestyle{empty}

\begin{abstract}
Large scale pre-training on text and image data along with diverse robot demonstrations has helped Vision Language Action models (VLAs) to generalize to novel tasks,  objects and scenes. However, these models are still susceptible to failure in the presence of execution-time impediments such as distractors and physical obstructions in the robot's workspace. Existing policy improvement methods finetune base VLAs to improve generalization, yet they still struggle in unseen distractor settings. 
To address this problem, we investigate whether internet-scale pretraining of large vision-language models (VLMs) can be leveraged to reason about these impediments and mitigate policy failures. 
To this end, we propose \textit{StageCraft}, a \emph{training-free} approach to improve pretrained VLA policy performance by manipulating the environment's initial state using VLM-based in-context reasoning. 
\textit{StageCraft} takes policy rollout videos and success labels as input and leverages VLM’s reasoning ability to infer which objects in the initial state need to be manipulated to avoid anticipated execution failures. \textit{StageCraft} is an extensible plug-and-play module that does not introduce additional constraints on the underlying policy, and only requires a few policy rollouts to work. We evaluate performance of state-of-the-art VLA models with \textit{StageCraft} and show an absolute $40\%$ performance improvement across three real world task domains involving diverse distractors and obstructions. 
Our simulation experiments in RLBench empirically show that \textit{StageCraft} tailors 
its extent of intervention based on the strength of the underlying policy and improves its performance with more in-context samples. Videos of \textit{StageCraft} in effect can be found at \href{https://stagecraft-decorator.github.io/stagecraft/}{\textcolor{LinkPink}{https://stagecraft-decorator.github.io/stagecraft}}. 

\end{abstract}
\section{Introduction}
Vision Language Action (VLA) \cite{shukor2025smolvla,intelligence2025pi} models have seen rapid adoption in the field of robotics due to the promise of impressive generalization abilities. VLAs are often initialized from VLMs trained on internet-scale image and text data, and undergo pretraining on large number of demonstrations with diverse scenes, objects, embodiments~\cite{rath2025xmop} and skills, making them generalize better to unseen tasks~\cite{intelligence2025pi}. Despite this, current state-of-the-art VLAs do not work zero-shot for novel task settings and require further fine-tuning on task specific dataset collected in the downstream environment. This fine-tuning often results in catastrophic forgetting in VLAs \cite{hancock2025actions} impacting their generalization to novel objects and scenes. Moreover, the downstream datasets are relatively smaller in size and do not cover a wide distribution for the policy to generalize over novel objects or obstructions. This makes fine-tuned VLAs brittle to distractors in downstream tasks despite being pretrained on varied scenes and objects. 

To compensate for lack of distributional coverage in supervised learning~\cite{pumacay2024colosseum}, and in lieu of more accessible and foundational policies such as VLAs~\cite{shukor2025smolvla}, several methods have been proposed to improve policy performance on downstream tasks. Policy improvement methods refine the actions generated from the policy either by using the policy rollouts in the downstream environment~\cite{ren2024dppo} or by using a separately trained value function~\cite{nakamoto2025vgps}. However, these methods are expensive to deploy in the real world, and they still suffer from the persistent challenge of covering a sufficiently broad distribution in the downstream environment for the policy to generalize to novel distractors.\cite{pumacay2024colosseum}. Hence, in this work, we present \textit{StageCraft}, a complementary approach that improves policy performance by modifying the environment’s initial state rather than fine-tuning or augmenting the policy’s parameters. \textit{StageCraft} achieves this by physically manipulating objects in the scene and removing those that are anticipated to hinder successful task completion.

As VLA models gain popularity, the skills learned by these policies will be deployed in the presence of a wide variety of objects at test time. For example, a policy trained for cooking may encounter countless everyday objects on the kitchen countertop that were not present in its training distribution. It is unrealistic to continually collect more data and repeatedly fine-tune VLAs for every new deployment setting simply to maintain performance. Hence, we argue that reliable execution of pre-trained VLAs will require some form of pre-execution intervention that makes the domain more favorable for policy rollout. In this work, we take a first step in this direction by performing object-removal interventions that bring the scene back toward a visual distribution that is less susceptible to policy failure.


An important question that arises here is: how do we determine which distractors in the environment adversely affect the policy’s performance on the downstream task? Critically, we should not make any assumptions on the working of underlying pre-trained policy, nor the data that was used to train it. We also argue that naively manipulating the environment to remove all supposed distractor objects would be wasteful, especially so if the policy already generalizes well. For instance, a policy performing well in the presence of a specific distractor obviates the need for its removal. In other words, if a VLA is sophisticated enough to function well in the presence of distractors, it should not require any intervention. In contrast, if the policy fails due to distractors, minimal environment modifications are needed to improve downstream performance. 

To reason about unseen distractors at test time with black box policies, we leverage the in-context reasoning capabilities of VLMs~\cite{li2025survey}. Our approach prompts VLMs with a few rollout episodes of the policy in the downstream environment with \textit{expected} distractor objects.  We then leverage VLM's grounded text and vision based priors that help it reason on the environment states of these rollouts and the consequent success or failure of the policy. From the in-context episodes and the execution results, the VLM reasons about a new initial environment state, identifying objects in the scene that need to be physically manipulated for increasing the chance of a successful policy execution. This approach not only prevents us from making assumptions on the training data of the policy, but also allows us to record the policy's robustness to different objects.

Our method is inspired by how humans prepare their environment before executing a skill. Kirsh et al. \cite{kirsh1995intelligent} argue that experts require less deliberation about the environment and are robust to its varying configurations, executing their skills with little to no preparation. In contrast, non-experts can benefit from task environments that are prepared and more constrained. We use this insight to argue that as VLAs become more accomplished in executing their skill in diverse environments, they will require less intervention or preparation of the environment. Hence, \textit{StageCraft} is designed to first reason if manipulation of a distractor is at all necessary given the history of policy rollouts, and then to prepare the environment when doing so is deemed essential for successful policy execution.
Our contributions are as follows:  
\begin{itemize}
    \item We propose \textit{StageCraft} (shown in Fig. \ref{fig:banner}), a training-free method that reasons about the initial environment state based on the past history of policy rollouts, and manipulates only the distractor objects
    deemed necessary for improving policy performance.
    
    \item We experimentally evaluate \textit{StageCraft} in a multi-task setting on a Franka FR3 robot across three real-world task domains, and find that it improves the performance of SmolVLA \cite{shukor2025smolvla} and Pi0.5 \cite{intelligence2025pi} by an absolute average $40\%$ under diverse obstruction and clutter scenarios across the three real world task domains. 
    \item Our RLBench simulation results show that \textit{StageCraft} consistently improves performance as the number of in-context episodes increase, generalizes across base policies of varying strength, and remains effective under our ablation studies.
    \item We design \textit{StageCraft} to be an extensible plug-and-play module that can be used as a decorator for VLA execution and open-source the implementation so that the broader community can benefit from our research on initial environment state manipulation for policy performance improvement.
\end{itemize}

\section{Related Works}
\subsection{Generalization in VLAs and Policy Improvement}
While new VLAs are increasingly being proposed in literature, they still lack the generalization required for zero-shot usage in novel tasks. As a consequence, VLAs are often fine-tuned on narrow task-specific datasets, resulting in catastrophic forgetting of the generalization abilities gained during the pretraining phase \cite{hancock2025actions}. Further, VLAs are often evaluated in very limited settings \cite{fang2025intention} that closely match their training distribution. For instance, several works evaluate VLAs and note that they fall short in terms of execution with out-of-distribution objects \cite{fang2025intention, patil2025factorizing} and do not generalize well with complex language commands \cite{fang2025intention}.  

Policy improvement typically aims to improve the performance of pre-trained policies for downstream tasks using training methods such as reinforcement learning\cite{ren2024dppo}, adapter learning\cite{wagenmaker2025steering}, and value function guidance\cite{nakamoto2025vgps}. These methods are non-trivial to train, and often require several hundreds of episodic samples to improve policy performance for a given reward. In contrast to these methods, we take a training-free approach of improving the underlying policy \cite{patil2026you}, where we modify the initial state of the environment to avoid expected failures. Our plug-and-play method can also be used alongside any of these aforementioned policy improvement methods yielding further gains in performance. Moreover, our method \textit{conservatively} handles out of distribution objects not seen even during the training or policy improvement process, by using the reasoning capabilities of a VLM trained on internet scaled image and text data.

\subsection{Reasoning using VLM Models}
Vision Language Models (VLMs) have been leveraged in robotics for reasoning over the visual modality such as behavior-critics for robot trajectories \cite{guan2024task}, success detectors or reward models for tasks \cite{du2023vision}, and reasoning over manipulation failures \cite{duan2024aha}. Duan et al. \cite{duan2024aha} fine-tune a VLM on simulated failure data to develop detection and reasoning capabilities for manipulation failures. Interestingly, Ma et al. \cite{ma2024vision} show that VLMs can act as in-context value learners, and used this capability for task progress prediction. Our work uses the in-context learning capabilities of VLMs to reason about the initial environment state, given a history of rollout episodes and the resulting success or failure.

Several works have used reasoning mechanisms such as chain-of-thought (CoT) \cite{wei2022chain} in VLAs to improve generalization. CoT has been applied to text \cite{zawalski2024robotic}, image~\cite{zhao2025cot} and action~\cite{zhong2026acot} modalities in VLAs during the pre-training phase to improve generalization. However, these methods require extensive architectural changes \cite{zawalski2024robotic,zhong2026acot, zhao2025cot}, data re-annotation \cite{zawalski2024robotic}, and do not transfer to other VLA models. Moreover, these methods primarily operate through sub-task \cite{zawalski2024robotic} or sub-goal \cite{zhao2025cot,zhong2026acot} generation to improve the linguistic and visual generalization capabilities, and do not specifically reason about the environment preconditions before executing the skill. In contrast, \textit{StageCraft} is a plug-and-play module that leverages CoT and can be used for any downstream VLA before executing the policy in the environment. To the best of the authors’ knowledge, none of the existing methods \textit{reason} over policy performance to make environment interventions that improve policy success.

\subsection{Environment Preconditions}
Environment preconditions have traditionally been studied in planning domains where a skill is executed when the specified symbolic predicates are true for a given state \cite{kroemer2021review}. Sharma et al. \cite{sharma2020relational} learn a classifier on continuous representations for pairwise objects in the scene to predict if the precondition is satisfied or not. In contrast, \textit{StageCraft} reasons only over distractors in the scene, without explicitly modeling objects’ spatial positions. It also does not assume access to the policy’s training data, and generalizes conservatively to unseen objects. A closely related work is ReSET \cite{dai2025prepare}, which reconfigures the initial environment state to lie in a more concentrated distribution termed “anchor states.” While it addresses a different problem, ReSET assumes access to the policy training data and relies on additional human task-specific data, including scene resets and robot play data, to learn corrective actions. In contrast, \textit{StageCraft} uses a VLM as a high-level planner over primitive actions that manipulate the environment, without collecting extra human intervention data. Moreover, \textit{StageCraft} generalizes to unseen distractors and obstructions, unlike ReSET.


\begin{figure*}[t]
    \centering
    \includegraphics[width=0.8\textwidth]{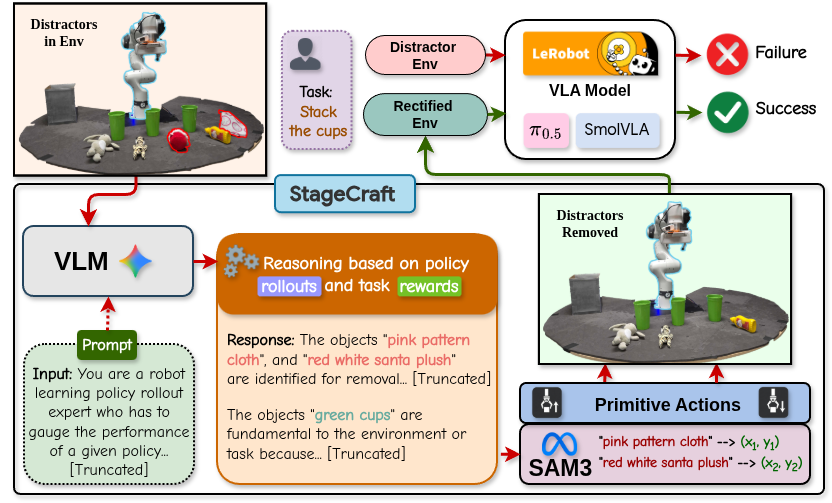}
    \caption{We empirically evaluate \textit{StageCraft} with two state-of-the-art VLA models, Pi0.5 \cite{intelligence2025pi} and SmolVLA \cite{shukor2025smolvla}, under challenging clutter and obstruction settings. In the setup shown, \textit{StageCraft} identifies \texttt{``pink pattern cloth''} and \texttt{``red write santa plush''} as potential failure-inducing distractors for the \textbf{\texttt{stack\_cups}} task. These objects are then detected and removed from the robot’s workspace via primitive pick-and-place actions, after which the policy is executed and successfully completes the task.}
    \label{fig:method}
\end{figure*}

\section{Method}
\label{sec:method}
\subsection{Problem Formulation}
We model a downstream task as a Markov decision process
$\mathcal{M} = (\mathcal{S}, \mathcal{O}, \mathcal{A}, p, H, \rho_0)$,
where $\mathcal{S}$ is the state space,
$\mathcal{O}$ is the observation space,
$\mathcal{A}$ is the action space,
$p(s_{t+1}\mid s_t,a_t)$ is the transition kernel,
$H$ is the horizon, and
$\rho_0$ is the initial-state distribution.
For notational convenience, we assume that an initial state
$s_0 \in \mathcal{S}$ admits a decomposition
$s_0 = (u_0, \mathcal{X}_0)$,
where $u_0$ denotes non-object factors such as robot configuration and environment parameters,
and $\mathcal{X}_0$ is the set of all objects present in the scene.
We further decompose
$\mathcal{X}_0 = \mathcal{X}_{\text{rev}} \cup \mathcal{X}_{\text{dist}}$,
where $\mathcal{X}_{\text{rev}}$ denotes a fixed set of \emph{task-relevant} objects that are present in every episode,
and $\mathcal{X}_{\text{dist}} \subseteq \mathcal{X}$ denotes the set of distractor objects.
Here $\mathcal{X} = \{x_1,\dots,x_N\}$ denotes the finite set of distractor objects that are expected to interact with the policy environment,
with $|\mathcal{X}_{\text{dist}}| \leq N$.
Each object $x_j$ is represented by its semantic category and pose.

We consider a fixed policy $\pi(a\mid o)$ that maps observations $o \in \mathcal{O}$ to actions.
We do not assume access to the policy’s training data or the MDP under which it was trained.
When deployed in the downstream environment $\mathcal{M}$ and initialized at
$s_0 = (u_0, \mathcal{X}_0)$, the policy may fail due to the presence of objects in
$\mathcal{X}_0$ that were not encountered during training.

Our goal is to develop a module, \textit{StageCraft}, that operates prior to policy execution
and infers a subset of objects
$\mathcal{X}_{\text{manip}} \subseteq \mathcal{X}_{dist}$
that should be manipulated. Importantly, \textit{StageCraft} should try to minimize the number of objects being manipulated $|\mathcal{X}_{\text{manip}}|$, with the upper bound being $\mathcal{X}_{\text{manip}} = \mathcal{X}_{\text{dist}}$. 
Applying the corresponding action primitives to these objects transforms the initial state to $s_0' = (u_0, \mathcal{X}_0')$ where $\mathcal{X}_0' = \{\mathcal{X}_{\text{rev}} \cup \mathcal{X}_{dist}\} \setminus \mathcal{X}_{\text{manip}}$.
The policy $\pi$ is then executed from the modified state $s_0'$ under the same environment dynamics.

Moreover, we assume access to a finite set of action primitives
$\mathcal{P} = \{\phi_1,\dots,\phi_K\}$,
which can be applied to transform the initial state $s_0$ to a modified state $s_0'$.
Each primitive $\phi_k$ is a parameterized, open-loop controller that operates on the environment state
(e.g., grasping, go-to-point, pick, place). 
Execution of a primitive induces state transitions according to the same environment dynamics $p$.

\subsection{StageCraft: Object-Set Creation}
\label{sec:stagecraft_creation}
For a given initial state $s_0$, \textit{StageCraft} infers a subset of distractor objects,
$\mathcal{X}_{\text{manip}} \subseteq \mathcal{X}_{\text{dist}}$,
whose removal can maximize the policy's success rate. Additionally, \textit{StageCraft} aims to minimize the number of removed distractors $|\mathcal{X}_{\text{manip}}|$ whenever the policy’s success rate is not anticipated to decrease. 

There are two challenges in this problem: (i) We do not know which objects pose as distractors for the policy, as we do not assume access to the policy’s training data or environment. Defining object relevance for the policy requires a causal understanding of how the policy was trained, which is harder to reason about without knowing the training distribution or goals of a task. Hence,  \textit{StageCraft} does not explicitly assume knowledge of the relevant objects in the scene. Instead, \emph{\textit{StageCraft} treats all objects in the scene on equal footing and operates over sets of objects that empirically demonstrate high performance across episodes.} Even if \textit{StageCraft} was aware of the objects in the training distribution, naively removing \emph{all} distractor objects from the scene, especially for VLAs pretrained on varied objects, skills and scenes is wasteful and unnecessary. Hence, \emph{\textit{StageCraft} uses a few policy rollouts as a weak Monte Carlo estimate of the underlying policy’s robustness to different sets of objects from collected episodes.}

To estimate the policy’s robustness to distractors, in a manner akin to online policy improvement, we collect rollouts of the fixed (VLA) policy $\pi$ in the MDP $\mathcal{M}$ from different initial states $s_0$, each instantiated with a distractor set $\mathcal{X}_{\text{dist}} \subseteq \mathcal{X}$.
For each fixed distractor subset $\mathcal{X}_{\text{dist}}^{(i)}$ among the $N$ total sets, we collect $M$ episodes by varying the object poses to obtain a reliable estimate of the policy’s success rate. Emphasizing the object-centric structure of the initial state:
\[
\mathcal{B}
=
\left\{
\big(u_0^{(i,j)}, \mathcal{X}_0^{(i)}, y^{(i,j)}\big)
\right\}_{i=1,\;j=1}^{N,\;M},
\]
where $\mathcal{X}_0^{(i)}$ denotes the $i$-th object set,
$u_0^{(i,j)}$ corresponds to the $j$-th instantiation (e.g., pose configuration) of that set,
and $y^{(i,j)} \in \{0,1\}$ denotes the success of the corresponding rollout. A small value of $M$ leads to a poor Monte Carlo estimate of the policy’s success rate for each distractor configuration,
while a small value of $N$ limits \textit{StageCraft}'s ability to estimate the policy’s robustness to different objects,
potentially causing it to default to removing all distractors, i.e.,
$\mathcal{X}_{\text{manip}} = \mathcal{X}_{\text{dist}}$.

\subsection{StageCraft: Object-Set Transition}
\label{sec:stagecraft_transition}
For each distractor set $\mathcal{X}_{\text{dist}}^{(i)}$, an empirical success rate $\mathrm{sr}^{(i)}$ is computed from the collected rollouts $\mathcal{B}$.
We retain a collection of distractor subsets
\[
\mathcal{S} = \left\{ s_j \subseteq \mathcal{X} \;\middle|\;
\mathrm{sr}(s_j) \geq \mathrm{sr}_{\max} \right\},
\]

where $\mathrm{sr}_{\max} = \max_{s \subseteq \mathcal{X}} \mathrm{sr}(s)$ denotes the highest observed success rate among all in-context episodes.

Given a new initial state
$s_0 = (u_0, \mathcal{X}_{\text{rev}} \cup \mathcal{X}_{\text{dist}})$,
the largest subset $s_j \in \mathcal{S}$ is selected such that
$s_j \subseteq \mathcal{X}_{\text{dist}}$.
The difference between $\mathcal{X}_{\text{dist}}$ and $s_j$
determines the set of object-level modifications to be executed by the action primitives
in order to transform the environment prior to policy execution. 
In other words, among the subsets of objects in $\mathcal{S}$ empirically observed to result in the highest success rate, we choose the largest subset that has all the objects present in the new scene $\mathcal{X}_0$. All the remaining objects in $\mathcal{X}_0$ form $\mathcal{X}_{manip}$.

This construction also ensures that \textit{StageCraft} acts conservatively with respect to policy performance in the presence of distractors.
Unseen distractors are automatically removed, since no subset in $\mathcal{S}$ contains them.
If the underlying policy is robust to distractors, then with sufficient rollouts,
$\mathcal{S}$ will include larger subsets containing multiple distractors,
providing stronger fallbacks in which policy performance is preserved.
Conversely, for policies that are not robust to distractors,
$\mathcal{S}$ will consist of subsets with fewer distractors,
requiring \textit{StageCraft} to apply more extensive object manipulations
prior to policy execution.
Overall, \textit{StageCraft} adapts its corrective actions to the robustness of the underlying policy,
scaling the degree of intervention according to the observed performance.

\begin{figure*}[t]
    \centering
    \includegraphics[width=0.9\textwidth]{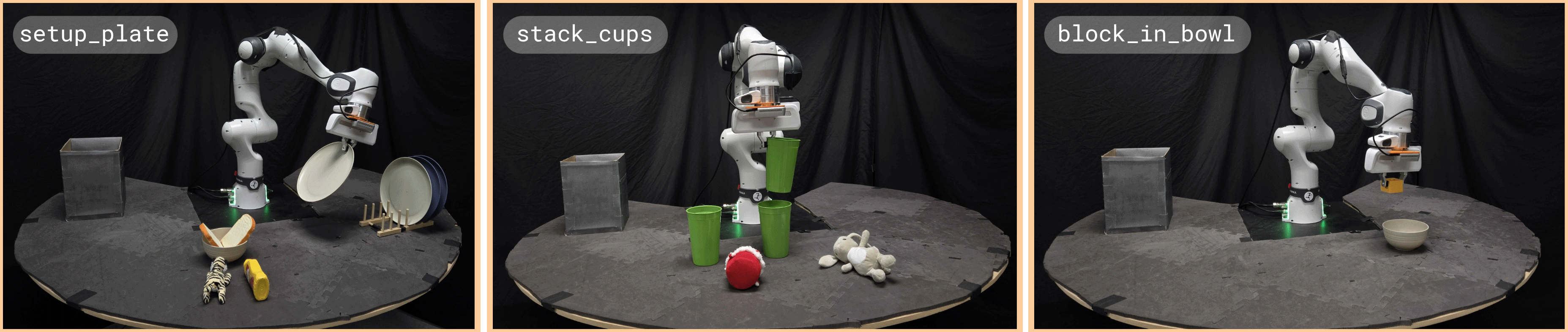}
    \caption{Real-world task domains used to evaluate \textbf{\textit{StageCraft}}. The \texttt{stack\_cups} and \texttt{setup\_plate} tasks require precise manipulation across four subtasks, whereas \texttt{block\_in\_bowl} is simpler and primarily requires accurate gripper alignment with the block to complete successfully. VLM reasoning followed by environment manipulation adds $\sim200$ seconds to the overall execution time.}
    \label{fig:tasks}
    
\end{figure*}

\subsection{Vision Language Model Reasoning}
\label{sec:vlm_reasoning}
To execute \textit{StageCraft} we need consistent object-centric representations of the initial states of the collected episodes and for the new initial state observed during policy evaluation. As shown in Fig. \ref{fig:method}, our method leverages the internet scaled reasoning capabilities of Vision Language Models (VLMs) for obtaining the object centric representations and inferring the minimal manipulation steps required, taking into consideration the action primitives available for the robot. Specifically, we leverage in-context reasoning within VLMs to make it follow the object-set creation and transition strategy laid out in Section \ref{sec:stagecraft_creation} \ref{sec:stagecraft_transition}. The VLM is prompted to first create object-sets with consistent names based on the in-context initial states of the episodes. Next the VLM infers success rate for each object-set based on the reward assigned to the corresponding episodes and filters the object-sets having the highest success rate. Once the object-sets have been filtered, the VLM executes the transition strategy, choosing the minimum manipulation actions to be executed in the given initial state while accounting for the skill affordances, leading to an object-set with the highest observed success rate. 
This strategy is explicitly specified to the VLM in the input language instruction prompt present on our webpage \href{https://stagecraft-decorator.github.io/stagecraft/}{\textcolor{LinkPink}{https://stagecraft-decorator.github.io/stagecraft}}.


We make two assumptions for \textit{StageCraft} that are necessary to guarantee an improvement over the baseline VLA. First, we ensure that the \textit{task-relevant} object set $\mathcal{X}_{\text{rev}}$ is present in at least a few successful policy rollouts used for the VLM in-context query. Second, we expect the VLM follows the set-transition strategy accurately as specified in the input language prompt. 
Under these assumptions, \textit{StageCraft} never removes any \textit{task-relevant} object that is required for task completion.

\subsection{Environment Modification using Primitive Actions}
The VLM generates short descriptions of the objects in the environment that should be removed from the robot’s workspace to avoid anticipated failures. We use these descriptions to prompt the SAM3 \cite{carion2025sam} model to detect bounding boxes for the corresponding objects. We then project the centers of these bounding boxes back into 3D coordinates in the robot’s base frame using calibrated camera extrinsics and intrinsics. Finally, we use an inverse kinematics based primitive motion planner to execute pick-and-place actions that remove the objects from the robot’s workspace and deposit them into a collection bin placed beside the robot.

\section{Results}
\begin{figure}[!t]
    \centering
    \includegraphics[width=0.45\textwidth]{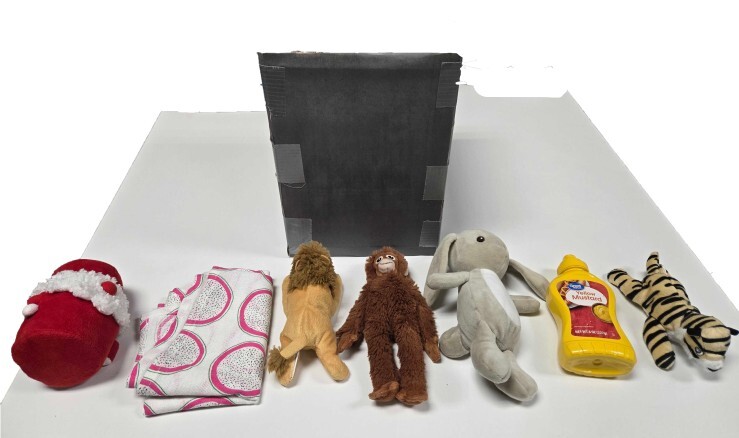}
    \caption{Set of $8$ distractor objects used in our real-world experiments. The gray collector bin is also a distractor as it was not present in the robot's workspace during data collection.}
    \label{fig:distractor_objects}
    
\end{figure}
We show \textit{StageCraft'}s ability to improve pretrained policy performance in downstream environments for Pi0.5 and SmolVLA through real robot and simulation experiments.

\subsection{Robot Experiment Setup}
\textit{StageCraft} was evaluated on three real-world task domains, as shown in Fig.~\ref{fig:tasks}. The tasks and their associated natural language instructions are as follows: \textbf{Stack Cups}, stack both the left and right cups on top of the center cup; \textbf{Setup Plate}, take a plate from the rack and serve a piece of bread on the plate; and \textbf{Block in Bowl}, pick up the block and place it in the bowl.
For VLA fine-tuning, we collected data for a larger set of $10$ tasks, which includes demonstrations for the three tasks above. We recorded $60$ demonstrations per task, and crowdsourced language instructions for each demonstration via Mechanical Turk \cite{crowston2012amazon}. We then fully fine-tuned the base VLA models, Pi0.5 and SmolVLA, using pretrained weights from LeRobot \cite{cadene2026lerobot}. Our dataset contains two camera streams from Intel D435 sensors: a wrist-mounted camera and a third-person front-facing camera. Proprioceptive state and actions are represented in absolute joint space. Our experiments include a fixed set of $8$ distractor objects, as shown in Fig.~\ref{fig:distractor_objects}. These objects were never present in the robot’s workspace during data collection, making them well-suited for testing visual out-of-distribution failures in the baseline VLA models. All our real-world experiments use \texttt{gemini-3.1-pro} VLM for reasoning. Additionally we show ablation results on \texttt{gemini-2.5-pro} and \texttt{gpt-5.2-pro} models. 

\subsection{Real-World Result Analysis}

\begin{figure}[t]
    \centering
    \includegraphics[width=0.5\textwidth]{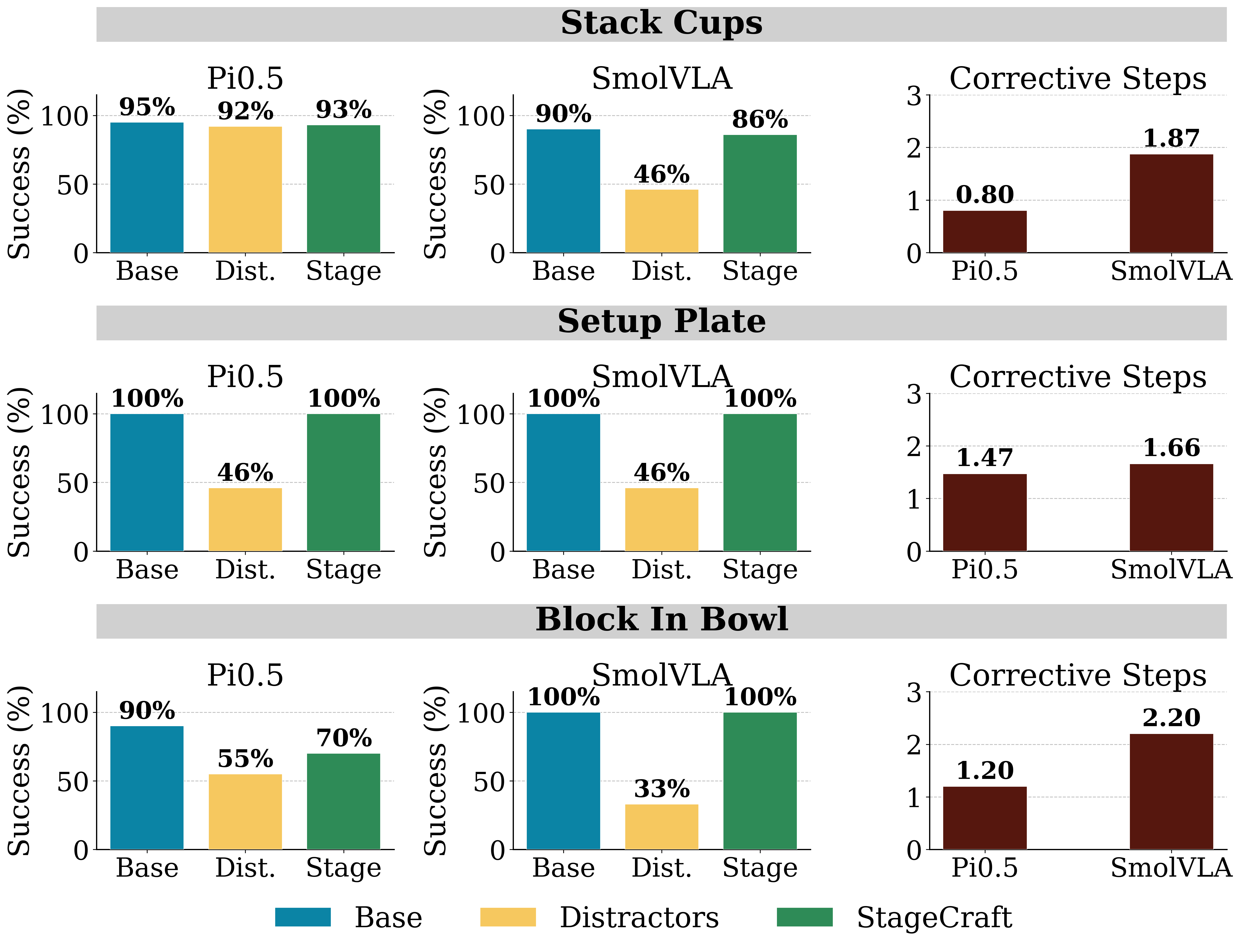}
    \caption{Task success rates for real-world robot experiments with Pi0.5 and SmolVLA under the \textbf{Base}, \textbf{Distractor}, and \textbf{StageCraft} settings. \textit{StageCraft} recovers baseline performance, bringing success rates closer to those observed in the \textbf{Base} setting. \textit{Stagecraft} intervenes with fewer corrective steps for the VLA with a stronger visual backbone, Pi0.5, than for the weaker SmolVLA, which is more vulnerable to visual distribution shifts.}
    \label{fig:real_robot_results}
    
\end{figure}

Fig.~\ref{fig:real_robot_results} shows the success rate (SR) of the baseline VLA models across three experimental settings: (1) \textbf{Base}, where the policy is evaluated in the same environment configuration used during data collection; (2) \textbf{Distractor}, where we introduce clutter and obstruction by adding 1--5 distractor objects to the base environment; and (3) \textbf{StageCraft}, where we modify the environment using our proposed method to improve the policy’s expected performance. Across all tasks, we observe consistent gains for both VLA baselines, with an average absolute SR improvement of $40\%$. We use $10$ policy rollouts for in-context reasoning and evaluate \textit{StageCraft} on $15$ additional rollouts in the \textbf{Distractor} setup. 

In our real-robot experiments, we ensured that all in-context episodes included the gray object-disposal bin in the robot’s workspace, since it is required to hide distractor objects from the camera view during the evaluation rollouts. However, this bin was absent during data collection. The reasoning traces demonstrate that \textit{StageCraft} consistently classifies the gray bin as a \textit{task-relevant} object across all real-world experiments, aligning with our minimal environment modification hypothesis.

\begin{figure}[t]
    \centering
    \includegraphics[width=0.5\textwidth]{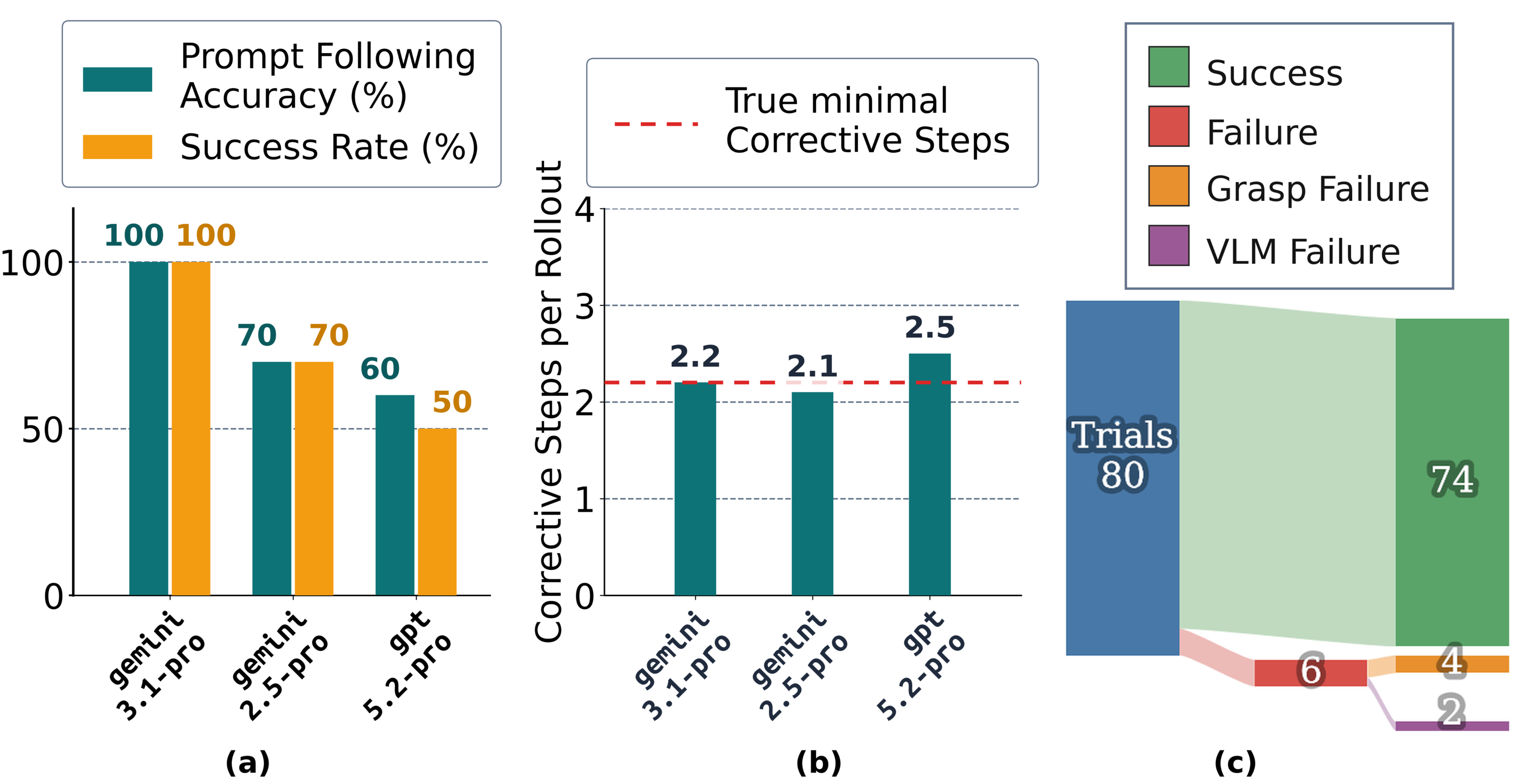}
    \caption{Comparison of performance of three different VLM's for (a)
    prompt following accuracy and task success rate, (b) corrective steps
    per rollout against the true minimal corrective steps, for the
    \texttt{block\_in\_bowl} task. Models with lower prompt following accuracy
    fail to take correct environment modification steps thus resulting in
    lower policy success rates. (c) Failure mode analysis
    across all three real-world tasks.
    }
    \label{fig:real_robot_ablation}
    
\end{figure}

Furthermore, we manually evaluate the reasoning trace for each rollout and compute prompt following accuracy by comparing the environment modification steps deduced by the VLM against the ground-truth. We find that \texttt{gemini-3.1-pro} VLM achieves an average $95\%$ prompt following accuracy across the three real-world tasks. A correct deduction requires the model to (i) consistently identify environment objects across all in-context observations, and (ii) correctly follow the set-transition prompt described in Section~\ref{sec:stagecraft_transition}. Overall, modern state-of-the-art VLMs were found to be more reliable on this task than older generations. 

Fig.~\ref{fig:real_robot_ablation} compares task success rate, prompt-following accuracy and the average number of environment modification steps taken by two additional VLMs, \texttt{gemini-2.5-pro} and \texttt{gpt-5.2-pro}, for the \texttt{block\_in\_bowl} task. We observe that older generation VLMs often fail to consistently identify the same objects across rollouts, and consequently struggle to follow the set-transition strategy correctly. Models with higher prompt-following accuracy naturally align more closely with the required minimal steps, supporting our hypothesis of performance-informed minimal environment modification as discussed in Section~\ref{sec:stagecraft_transition}. Based on this result, we expect future generations of VLMs to further improve prompt-following accuracy and reduce environment modifications within the \textit{StageCraft} framework.

\subsection{Simulation Experiments}

Our simulation experiments are designed to answer the following questions regarding \textit{StageCraft}'s capabilities: (i) Does \textit{StageCraft} use the underlying policy's robustness to distractors to inform the extent of environment interventions required? (ii) Does the performance of \textit{StageCraft} improve upon increasing the number of samples for each distractor setting? (iii) Does the object-set creation and transition strategy lead to more effective and reliable environment interventions than naively querying a VLM?

\begin{figure}[!t]
    \centering
    \includegraphics[width=\columnwidth]{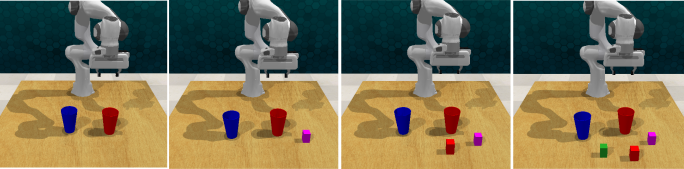}
    \caption{RLBench environments \texttt{Zero}, \texttt{One}, \texttt{Two}, \texttt{Three} in order from left to right, with the numeric name denoting the number of distractors in the scene. Environment \texttt{Three} is used for evaluating \textit{StageCraft}.}
    \label{fig:rlbench environments}
\end{figure}

We investigate these questions by training two SmolVLA~\cite{shukor2025smolvla} policies on the task \texttt{pick the red cup} using $50$ and $250$ demonstrations, referred to as $\pi_{\text{weak}}$ and $\pi_{\text{strong}}$, respectively. 
The resulting policies achieve success rates of $78\%$ for $\pi_{\text{weak}}$ and $95\%$ for $\pi_{\text{strong}}$ on the base environments named \texttt{Zero}. As illustrated in Fig. \ref{fig:rlbench environments}, we create three distractor variants of the environment, \texttt{One}, \texttt{Two} and \texttt{Three}. All experiments use \texttt{gemini-2.5-pro} as the VLM due to its 
relatively lower inference latency. We expect better performance with better VLMs as empirically shown in our real robot experiments.

\begin{figure}[!t]
    \centering
    \includegraphics[width=\columnwidth]{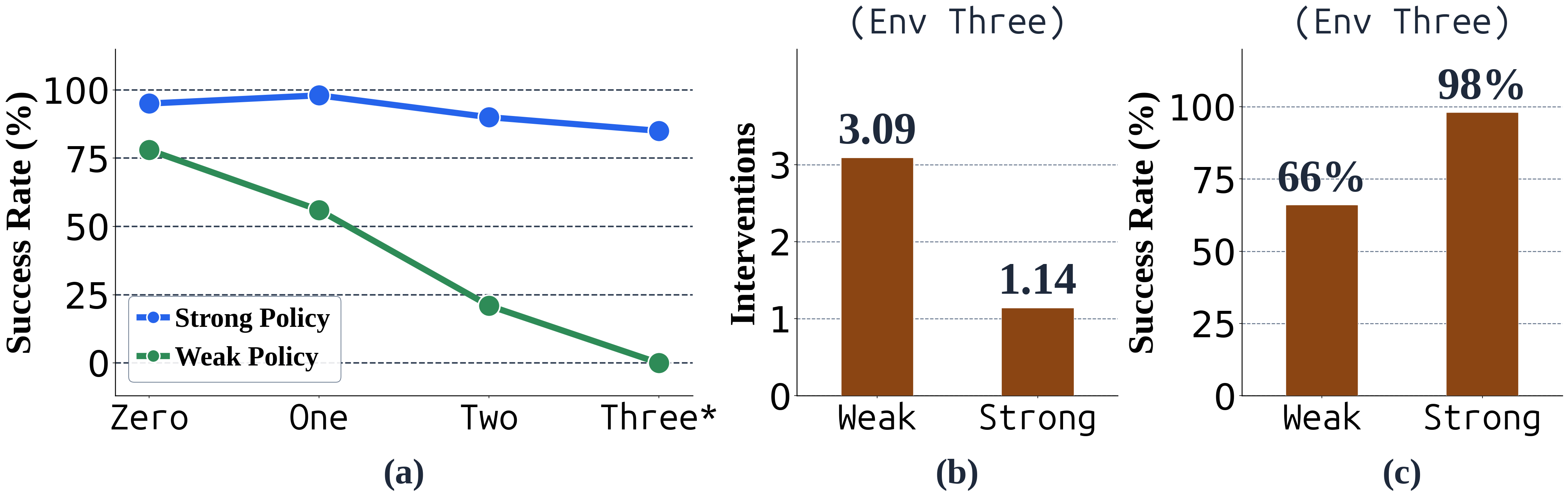}
    \caption{Improvements on Strong and Weak policies (a) Success rates for $\pi_{\text{weak}}$ and $\pi_{\text{strong}}$ on RLBench environments. We see that the performance of $\pi_{\text{weak}}$ deteriorates significantly as distractors are introduced, while that of $\pi_{\text{strong}}$ stays relatively strong. (b, c) This trend is also reflected in the number of interventions carried out by \textit{StageCraft} in environment \texttt{Three}, resulting in an improved success rate.}
    \label{fig:weak_strong_res}
    
\end{figure}

To evaluate how \textit{StageCraft} adjusts its extent of intervention based on the strength of the underlying policy, we evaluate it with $\pi_{\text{weak}}$ and $\pi_{\text{strong}}$ in the environment \texttt{Three}. We populate the in-context buffer with rollouts from \texttt{Zero}, \texttt{One} and \texttt{Two} for each policy, respectively. As seen in Fig. \ref{fig:weak_strong_res}, $\pi_{\text{strong}}$ experiences less deterioration in performance as distractors increase, whereas the performance of $\pi_{\text{weak}}$ swiftly drops. Accordingly, \textit{StageCraft} takes $3.09$ steps on average over $100$ evaluation episodes for $\pi_{\text{weak}}$, and only $1.14$ steps in the same setting for $\pi_{\text{strong}}$. \textit{StageCraft} improves the performance of $\pi_{\text{strong}}$ by $13\%$, increasing it to $98\%$, and improves the performance of $\pi_{\text{weak}}$ by $66\%$ from $0\%$ in the evaluation environment \texttt{Three}. These results demonstrate that \textit{StageCraft} adjusts its intervention based on the strength of the underlying policy, taking more corrective steps for weaker policies while yielding larger gains in performance.

\begin{figure}[!t]
    \centering
    \includegraphics[width=\columnwidth]{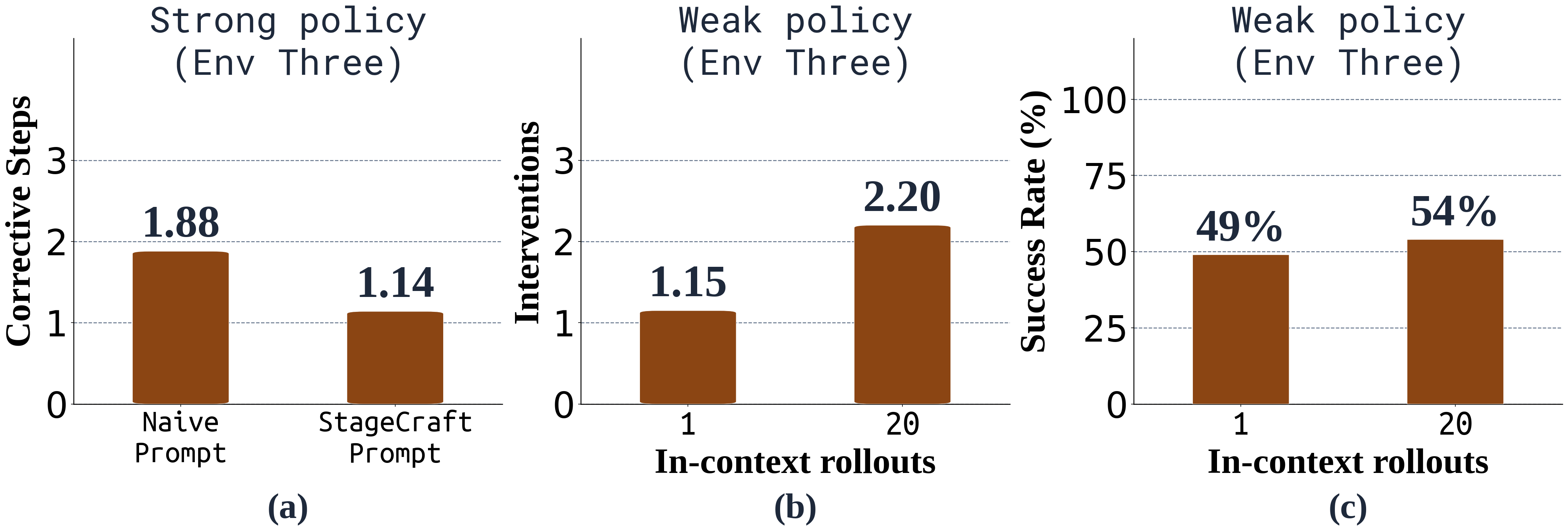}
    \caption{Importance of \textit{StageCraft} (a) \textit{StageCraft} prompt results in fewer corrective steps compared to naive-prompting. (b, c) More in-context episodes help estimate the required number of interventions more accurately while also improving policy performance.}
    \label{fig:icl_results}
    
\end{figure}

Next, we evaluate how \textit{StageCraft} performs with more in-context samples per distractor setting. Specifically, we evaluate whether \textit{StageCraft} benefits from a better Monte Carlo estimation of the policy's robustness to the distractors in an environment. We do this by evaluating the performance of $\pi_{\text{weak}}$ with \textit{StageCraft} when VLM's in-context buffer is populated with 1 and 20 episodes of both environments \texttt{One} and \texttt{Two}. The performance of $\pi_{\text{weak}}$ on the evaluation environment \texttt{Three} is shown in Fig. \ref{fig:icl_results} (b, c). We find that when only one episode for each of \texttt{One} and \texttt{Two} are included in \textit{StageCraft's} in-context buffer, its estimation of the policy's robustness to the distractors in the scene negatively affects it's performance. For example, when just one successful episode for environment \texttt{Two} is included in the VLM's context, \textit{StageCraft} assumes that the policy is robust to distractors in environment \texttt{Two} and does not act to remove the corresponding object set. However, this negatively affects the policy performance, as the policy is only achieves a success rate of $21\%$ in environment \texttt{Two}, as seen in Fig. \ref{fig:weak_strong_res} (a). Including more in-context samples for environment \texttt{Two} generates a better estimate of the policy performance with the corresponding distractors and helps \textit{StageCraft} take the required corrective steps. This can be seen from the increase in the number of average interventions ($2.2$ as compared to $1.15$) and the consequent increase in performance ($54\%$ as compared to $49\%$) with $20$ in-context rollouts for environments \texttt{One} and \texttt{Two} over just $1$ rollout of each.
Thus, the performance of our method is expected to improve with more in-context samples per distractor setting.

Finally, we evaluate whether our object-set creation and transition formulation is important for \textit{StageCraft}, and whether naively prompting the VLM yields the same result. Specifically, we exclude the object-set creation and transition strategy from the prompt to the VLM, while preserving the larger objective: to remove the minimum number of distractors from the scene to improve policy performance based on the provided in-context episodes. To test this, we evaluate $\pi_{\text{strong}}$ on a novel distractor environment with 10 in-context samples for each distractor setting. As shown in Fig.~\ref{fig:icl_results} (a) not only did the naive-prompt devoid of the object-set strategy (\texttt{prompt-baseline}) remove more objects on average ($1.88$ compared to $1.14$ for \textit{StageCraft}), but it also removed items crucial for task completion rather than distractors, such as the blue cup itself. Out of $25$ episodes that we qualitatively analyzed, \texttt{prompt-baseline} was less reliable in terms of the number of objects removed (with a coefficient of variance of $57.8\%$ compared to $13.62\%$). It also removed all distractors in 4 cases, and the blue cup itself in 12 cases. This highlights that the object-set based formulation leads to more consistent and accurate predictions, and relying solely on the open-ended reasoning capabilities of state-of-the-art VLMs does not yield consistent or accurate inferences about the initial environment state. 

\section{Conclusion}
We propose \textit{StageCraft}, a novel method that physically intervenes to manipulate the environment's initial state to improve the performance of VLA policies in the presence of distractors. \textit{StageCraft} informs its interventions based on the policy's robustness to sets of objects in the scene, estimated from the in-context episodes. Through extensive real-world experimentation, we show that we are able to improve the performance of Pi0.5 and SmolVLA by an absolute average of $40\%$ on three downstream tasks in the presence of distractors, with only $10$ in-context episodes. Through our simulation experiments, we show that \textit{StageCraft} adapts the extent of its intervention to the performance of the underlying policy, and identifies potential distractors more accurately as the number of in-context samples increases. We ablate our method to show the utility of the proposed object-set formulation and empirically demonstrate that open-ended reasoning using state-of-the-art VLMs generates inconsistent and inaccurate responses to improve policy performance in the presence of distractor objects. 

\textbf{Limitations}: The capabilities of \textit{StageCraft} are ultimately bounded by the underlying VLM. As we scale the number of rollouts to obtain better Monte Carlo estimates of VLA performance, we eventually hit the VLM’s context-length limit, since each episode includes image observations that are token-heavy. 
However, we do expect future VLMs with larger context windows to scale better in favor of \textit{StageCraft}. 
This approach would also have to be modified in cases where distractor objects are numerous or when it is impossible to manipulated the distractor itself. 
In this work, we do not address all possible distractor scenarios and only consider object-based sets. Consequently, \textit{StageCraft} is not suitable for environments in which distractors cannot be organized into discrete object sets.


\bibliographystyle{IEEEtran}
\bibliography{references}
\end{document}